\documentclass{ecai}  

\usepackage{graphicx}
\usepackage{latexsym}
\usepackage{caption}
\usepackage{subcaption}
\usepackage{xcolor}
\usepackage{amsmath,amsfonts,bm}
\usepackage{amsthm}
\usepackage{multirow}
\usepackage{booktabs}
\usepackage{algorithm}
\usepackage{algorithmic}
\usepackage[switch]{lineno}

\begin{document}

\begin{frontmatter}

\title{An Empirical Study of Retrieval-enhanced \\ Graph Neural Networks}



\author{\fnms{Dingmin}~\snm{Wang}$^1$, \fnms{Shengchao}~\snm{Liu}$^2\thanks{ Both Bernardo and Qi serve as the corresponding authors of this paper. Shengchao and Hanchen make equal contribution to this paper.}$, \fnms{Hanchen}~\snm{Wang}$^3$, \fnms{Bernardo}~\snm{Cuenca Grau}$^1$, \\\fnms{Linfeng}~\snm{Song}$^4$,  \fnms{Jian}~\snm{Tang}$^2$, \fnms{Le}~\snm{Song}$^5$, \fnms{Qi}~\snm{Liu}$^6$}

\address{$^1$University of Oxford, $^2$Mila-HEC Montreal, $^3$University of Cambridge, $^4$Tencent AI Lab}
\address{$^5$BioMap and MBZUAI, $^6$The University of Hong Kong}

\begin{abstract}
Graph Neural Networks~(GNNs) are effective tools for graph representation learning. Most GNNs rely on a recursive neighborhood aggregation scheme, named message passing, thereby their theoretical expressive power is limited to the first-order Weisfeiler-Lehman test (1-WL).  An effective approach to this challenge is to explicitly retrieve some annotated examples used to enhance GNN models.  While retrieval-enhanced models have been proved to be effective in many language and vision domains, it remains an open question how effective retrieval-enhanced GNNs are when applied to graph datasets. Motivated by this, we want to explore how the retrieval idea can help augment the useful information learned in the graph neural networks, and we design a retrieval-enhanced scheme called \textsc{GraphRetrieval}, which is agnostic to the choice of graph neural network models.
In \textsc{GraphRetrieval}, for each input graph, similar graphs together with their ground-true labels are retrieved from an existing database. Thus they can act as a potential enhancement to complete various graph property predictive tasks. We conduct comprehensive experiments over $13$ datasets, and we observe that \textsc{GraphRetrieval} is able to reach substantial improvements over existing GNNs. Moreover, our empirical study also illustrates that retrieval enhancement is a promising remedy for alleviating the long-tailed label distribution problem. 
\end{abstract}

\end{frontmatter}

\section{Introduction} \label{sec:intro}
Graph neural networks (GNNs) are a class of neural architectures
for supervised learning which has been adopted in a plethora of applications involving graph-structured data, such as molecular design
\cite{DBLP:conf/icml/GilmerSRVD17,DBLP:conf/nips/LiuABG18}, drug discovery~\cite{cheung2020graph}, recommendation systems \cite{DBLP:conf/kdd/YingHCEHL18,gao2022graph,wu2022graph,wang2019knowledge}, and knowledge graph completion \cite{DBLP:journals/corr/SchlichtkrullKB17,zhang2020relational}.

\begin{figure*}
     \centering
     \begin{subfigure}[b]{0.31\textwidth}
         \centering
         \includegraphics[width=\textwidth]{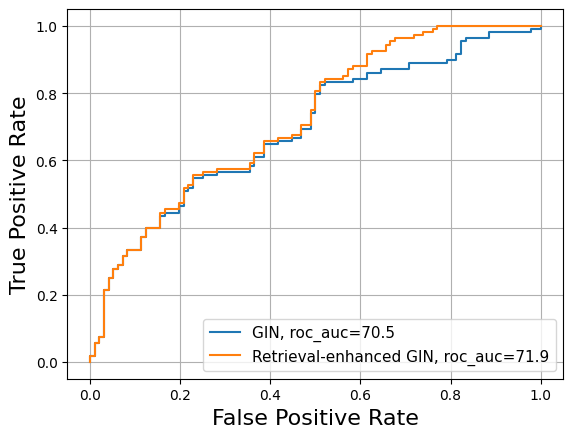}
         \caption{Binary Classification}
         \label{fig:y equals x}
     \end{subfigure}
     \hfill
     \begin{subfigure}[b]{0.31\textwidth}
         \centering
         \includegraphics[width=\textwidth]{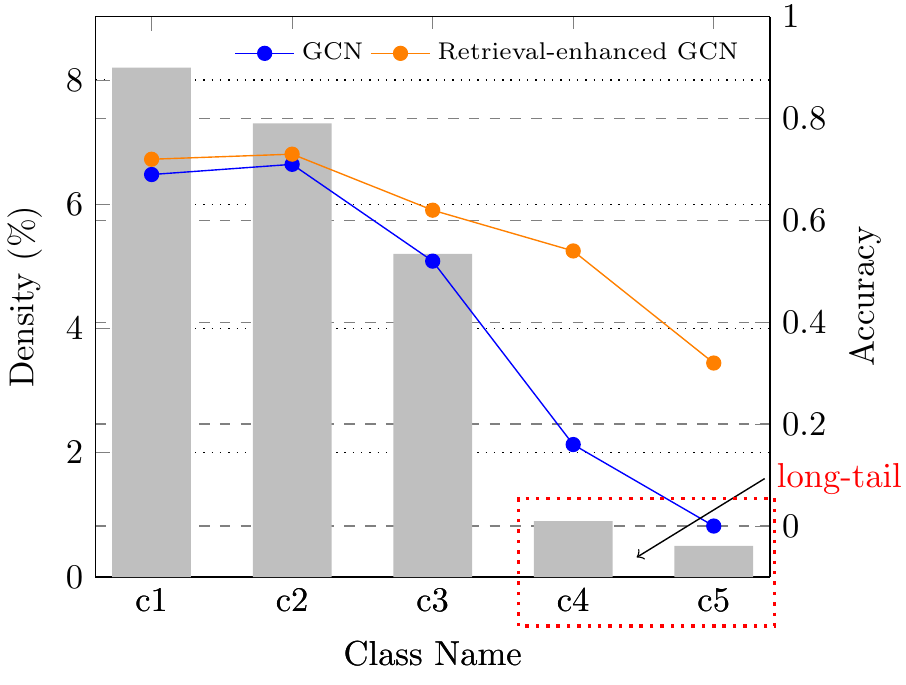}
         \caption{Multi-class Classification}
         \label{fig:three sin x}
     \end{subfigure}
     \hfill
     \begin{subfigure}[b]{0.31\textwidth}
         \centering
         \includegraphics[width=\textwidth]{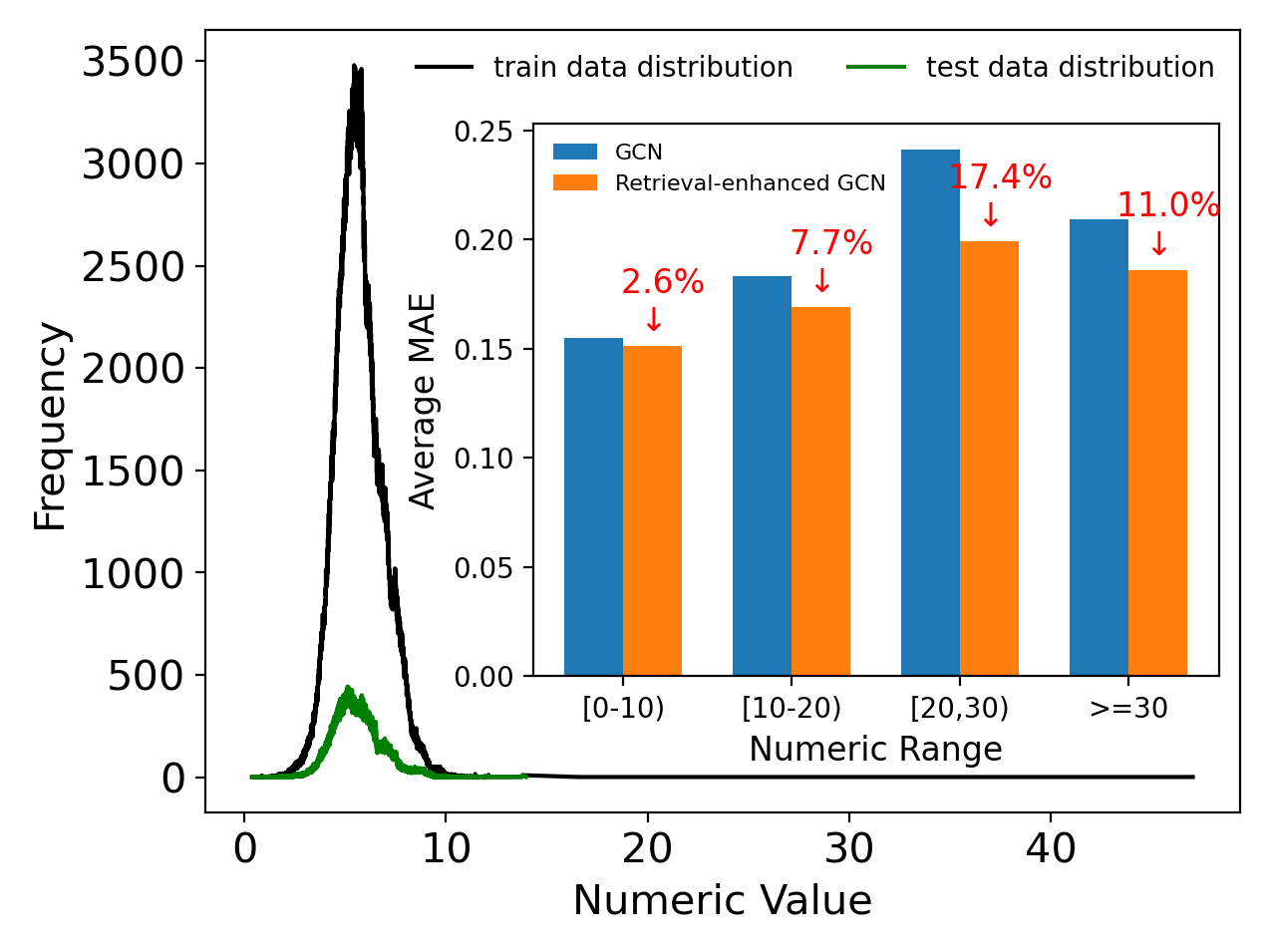}
         \caption{Regression}
         \label{fig:five over x}
     \end{subfigure}
        \caption{Retrieval-enhanced graph neural networks achieve consistent improvements over existing GNN models. \textbf{(a)} Experimental results of  GIN w/ and w/o retrieval-enhancement on an OGB dataset~(BBBP); \textbf{(b)} Part of experimental results of GCN w/ and w/o retrieval-enhancement on a reactant-product catalyst dataset with 888 common catalyst types (USPTO). The \textcolor{gray}{grey} bar represents the percentage of per-class samples in a class-imbalanced training dataset and \textcolor{blue}{blue} and \textcolor{orange}{yellow} line charts denote the per-class accuracy in a class-balanced testing dataset (100 samples per class); \textbf{(c)} The black 
 line and the green line represent the label distribution of the training and testing dataset of PCQM4M, respectively, and the bar chart represents the average MAE results of the testing datasets in each numeric range. The numeric ranges [10,20), [20,30) and $\geq 30$ are considered as long-tailed numerical ranges, where we see a much larger improvement than that in [0,10).}
        \label{fig:example}
\end{figure*}

To improve the representation power and performance of GNN models, existing studies mainly focus on the architecture design, and a growing plethora of new GNN models~\cite{kipf2016semi,xu2018powerful,corso2020principal} have been proposed in the past few years. Despite that newly-designed GNN models have brought improvement over a wide range of graph-based predictive tasks, 
some open challenges remain to be unsolved.
One such issue is the example forgetting a.k.a. catastrophic forgetting \cite{DBLP:journals/corr/abs-1812-05159,zhou2021overcoming,liu2021overcoming} since the training process does not ensure that the trained model will completely ``remember'' the training examples. 


Motivated by the success of retrieval-enhanced models in language and vision domains~\cite{lewis2020pre,guu2020retrieval,rao2021student}, in this paper, we aim at exploring the effects of retrieval-enhancement in the regime of GNNs. Our experimental results show that the performance of GNN models can be enhanced if the training data is also exploited at the testing time for making predictions. This is in contrast to the supervised learning setting, where the training data is
typically discarded once training has been completed
and the model relies solely on the learned parameters for making predictions with respect to  the given input. 

In this study, we focus on two common tasks in the realm of graph neural networks: graph classification and graph regression. Given training examples
consisting of graph-label pairs $(\mathcal{X}_{i},l_{i})$, 
where nodes in each graph are annotated with numeric feature vectors, we aim to learn a model that predicts label $l$ given an input $\mathcal{X}$.
At training time, \textsc{GraphRetrieval} consists of the following steps.
\begin{enumerate}
    \item \emph{Standard supervised training}. Initially, we train the GNN-based model of interest from 
    examples in the usual way, and obtain a trained model $\mathcal{M}$ as a result.
    
    \smallskip 
    
    \item \emph{Indexing.} We apply $\mathcal{M}$ to each of the training examples $\{(\mathcal{X}_1, l_1), \ldots,  (\mathcal{X}_n, l_n)\}$ and obtain the corresponding 
    representation vectors
    $h_{\mathcal{X}_1}, \ldots, h_{\mathcal{X}_n}$. We then 
    construct a single Maximum Inner Product Search (MIPS) index using FAISS~\cite{johnson2019billion}, where the key is  $h_{\mathcal{X}_i}$ and the value is the  corresponding example 
    $(\mathcal{X}_i, l_i)$.
    
    \smallskip 
       
    \item \emph{Self-attention based adapter}. The adapter is parameterized with two learnable matrices $\mathbf{W_1}$ and  $\mathbf{W_2}$ and a bias vector $\mathbf{b}$, which can be optimised using the training data after the model $\mathcal{M}$ and the index have been obtained. Note that $\mathbf{W_1}$, $\mathbf{W_2}$ and $\mathbf{b}$ are the only additional parameters required on top of the baseline GNN.
\end{enumerate}
Given an input 
$\mathcal{X}$ at testing time, we compute its
label $l$ according to the following steps.
\begin{enumerate}
\item \emph{Standard GNN application}. We apply $\mathcal{M}$ to $\mathcal{X}$ to obtain its representation vector $h_{\mathcal{X}}$
and its corresponding initial label $l_{\mathcal{X}}$ in the usual
way.
\item \emph{Graph retrieval}. We rank the representation vectors
in the index according 
to their L2 similarity to 
$h_{\mathcal{X}}$, and retrieve the top$-k$
vectors $\{h_{\mathcal{X}^{1}}, \ldots h_{\mathcal{X}^{k}}\}$ with highest similarity score; we then
retrieve the corresponding example graphs and labels  $(\mathcal{X}^{1}, l^{1}), \ldots (\mathcal{X}^{k}, l^{k})$.
\item \emph{Self-attention}. 
Based on the learned adapter and $\mathcal{M}$, we compute the final output label $l$ as a weighted sum of the initially predicted label  $l_{\mathcal{X}}$ and the labels  $l^{1}, \ldots l^{k}$ of the retrieved training examples.
\end{enumerate}

We have implemented our approach and 
used it to enhance three strong GNN backbone models, GCN \cite{kipf2016semi}, GIN \cite{xu2018powerful}, and PNA \cite{corso2020principal}. 
The effectiveness and feasibility of \textsc{GraphRetrieval} are evaluated on $13$ benchmarks, including twelve one-instance graph datasets and one multi-instance graph dataset. For example, in Figure~\ref{fig:example} (a), retrieval-enhanced GIN achieves a 2\% ROC\_AUC improvement on the BBBP dataset when compared with GIN without using the retrieved results; besides, we conduct experiments over two datasets with \textit{long-tailed label distribution}~\cite{zhang2017range,menon2020long,hong2021disentangling}, wherein many labels are associated with only a few samples.
From Figure~\ref{fig:example} (b) and (c), we can observe that our proposed approach brings consistent improvements over baselines in both the classification and the regression tasks. In particular, from the dissected experimental results, we can notice that the retrieval enhancement exhibits a significant improvement in the prediction of the long-tailed classes or values. Taking the bar chart in Figure~\ref{fig:example}(c) as an example, we can see only an average $2.6\%$ MAE improvement in the non-long-tailed numerical range but a more than $17\%$ improvement in the long-tailed number ranges. In summary, our contributions are as follows:
\begin{itemize}
    \item With negligible computational costs\footnote{On one hand, the computational cost of retrieving similar graphs using FAISS~\cite{johnson2019billion} is negligible due to the tool's well-proven efficiency and scalability. On the other hand, calculating the attention weights for the input graph and retrieved graphs in the adapter module only amounts to a single layer matrix operation. Therefore, the overall computational costs incurred by \textsc{GraphRetrieval} are trivial.}, we introduce \textsc{GraphRetrieval}, a simple yet effective framework for enhancing graph neural networks. \textsc{GraphRetrieval} is a flexible framework that can be applied to various graph prediction tasks, while remaining agnostic to the choice of graph neural network models.

    \item We extensively evaluate \textsc{GraphRetrieval} on thirteen datasets consisting of various graph sizes, ranging from small-scale single-instance molecular datasets to large-scale multi-instance therapeutics-related datasets. Our experimental results demonstrate that \textsc{GraphRetrieval} outperforms prior baseline models.

    \item Furthermore, our comprehensive empirical study of retrieval-enhanced graph neural networks demonstrates that they offer a promising approach to improving performance on datasets with long-tailed label distributions.
\end{itemize}
\section{Related Work} \label{sec:related}

In this section, we briefly discuss recent GNN architectures and retrieval-augmented models in Deep Learning.

\subsection{Graph Neural Networks}

Graph Neural Networks have become the 
de-facto standard for Machine Learning tasks over graph-structured data. The fundamental idea behind GNNs is that of \textit{Message Passing}---that is, to iteratively update each node-level representation by aggregating information from its neighborhoods throughout a fixed number of layers~\cite{kipf2016semi,gilmer2017neural,xu2018powerful,balcilar2021breaking}.

In molecule classification applications, 
Weave~\cite{kearnes2016molecular} explicitly learns both the atom- and bond-level representation during message passing;
D-MPNN~\cite{yang2019analyzing} emphasizes the information delivered along different directions; more recently, N-Gram Graph~\cite{liu2018n} and AWARE~\cite{demirel2021analysis} focus on modeling the walk-level information on molecules. The choice of the aggregation function is another key aspect of GNN design. GAT~\cite{velivckovic2017graph} adds an attention module; GGNN~\cite{li2015gated} combines information from its neighbors and its own representation with a GRU unit~\cite{cho2014properties}; finally, PNA~\cite{corso2020principal} adopts multiple aggregation functions for combining messages.

\subsection{Retrieval-augmented Models}

Prior studies in different domains show that models depending only on input features and learned parameters are less powerful than models augmented by external knowledge ~\cite{lewis2020pre,guu2020retrieval,wang2021template,kossen2021self,wu2022memorizing}. The authors of
\cite{wu2022memorizing} propose the use of a kNN-augmented attention layer for language modeling tasks.
In  \cite{kossen2021self}, the authors advocate for the use of the entire dataset (rather than using just a a single datapoint) for making predictions; in particular, they show that there exist meaningful dependencies between the input and the training dataset which are lost during training. Other related works such as retREALM~\cite{guu2020retrieval} and KSTER~\cite{jiang2021learning} augment the Transformer model by retrieving relevant contexts in a non-parametric way. 

In contrast to prior work, however, \textsc{GraphRetrieval} focuses on the retrieval of graph-structured data. Retrieving similar graphs can be much more challenging than retrieving similar text, where pre-trained models like BERT~\cite{devlin2018bert} can provide universal representations for measuring similarities. Such pre-trained models are missing for diverse graph-structured data. Furthermore, graph retrieval is also negatively affected by over-smoothing and over-squashing in existing GNNs, which are two unsolved and challenging issues inherent to current GNN models that might affect the learning of graph embeddings, which in turn may negatively impact the quality of retrieval.  Given that our framework allows for off-the-shelf use of \textit{flexible} GNN as the base model, so at the current stage, these issues are perpendicular to our proposed retrieval algorithm. However, we believe that our framework could achieve more gains in improving the quality of the predictions if GNNs without over-smoothing and over-squashing issues appear in the future.

\section{Background}\label{sec:background}

In this section, we recapitulate the basics of GNNs and formulate the graph 
classification and regression tasks accordingly.

\subsection{Message-passing GNNs}
In the context of GNNs, we define a graph
$\mathcal{G}$ as
a tuple 
$$\mathcal{G} := (\mathcal{V}, \mathcal{E}, \{\mathbf{x_v}\}_{v \in V}, \{\mathbf{x_{e}}\}_{e \in E}),$$
where $\mathcal{V}$ is a set of nodes
with 
each node $v \in V$ annotated with a numeric feature vector $\mathbf{x_v}$, and 
$\mathcal{E}$ is a set of (undirected)
edges where each edge $e \in E$ is also
annotated with a feature vector $\mathbf{x_e}$.
For instance, in a molecule classification
scenario, nodes may represent atoms, edges may represent bonds between molecules,
and feature vectors encode information about atoms and bond types.

 When given a graph $\mathcal{G}$ as input, a message-passing GNN~\cite{gilmer2017neural} with $T$
 layers updates each
 node feature vector
 throughout by aggregating messages from its neighbors. Formally,
 for each layer $1 \leq t \leq T$ and
 each node $v \in V$, the feature vector
 $\bm{h}_v^{(t)}$ for node $v$ in layer $t$
 is computed as follows:
\begin{equation}
    \bm{m}_v^{(t)} = \sum_{\{u,v\} \in E } M_{t-1} \big( \bm{h}_v^{(t-1)}, \bm{h}_u^{(t-1)}, x_{\{u,v\}} \big),
\end{equation}
\begin{equation}
    \bm{h}_v^{(t)} = U_t \big(\bm{h}_v^{(t-1)}, \bm{m}_v^{(t)} \big),
\end{equation}
where  $\bm{h}_v^{(0)} = \bm{x}_v$ for each $v \in V$, and where
$M_t$ and $U_t$ 
represent the GNN's message passing and update functions in the $t$-th layer,
respectively. The concrete description of these functions is given in terms of the learnable parameters of the model. Finally, a ReadOut function is applied to the node feature vectors
in the outermost layer $t=T$ to obtain a graph-level representation vector:
\begin{equation}\label{eq:graph}
    \bm{h}_\mathcal{G} = \textbf{AGG}\bigg(\bigg\{\bm{h}_v^{(T)} \mid  v \in \mathcal{G}\bigg\}\bigg),
\end{equation}
where  $\textbf{AGG}$ can be any permutation-invariant 
operation on multi-sets applied to a vector
in a component-wise manner, such as average, summation, or attention-based aggregation.

\subsection{Graph Property Prediction} 
Given an input $\mathcal{X}$, there are usually two tasks of interest:
\begin{enumerate}
    \item \textit{Single-instance prediction}, where $\mathcal{X}$ is a single graph and the target is to classify $\mathcal{X}$ into one of predefined classes or map $\mathcal{X}$ to a numeric value.
    \item \textit{Multi-instance prediction}, where $\mathcal{X}$ usually consists of two or more graphs and the target is also mapping $\mathcal{X}$ to  one of predefined classes or a numeric value. 
\end{enumerate}

For both \textit{Single-instance prediction} and \textit{Multi-instance prediction}, a basic structure could be formulated as follows:
\begin{equation}
    \bm{h}_\mathcal{X} = \bigoplus_{\mathcal{G} \in \mathcal{X}} \text{GNN}(\mathcal{G})
\end{equation}
\begin{equation}\label{eq:binary}
    l_{\mathcal{X}} = \Phi(\bm{h}_\mathcal{X}),
\end{equation}
where $\bigoplus$ is a concatenation operation,  $\bm{h}_\mathcal{X}$ denotes the representation of the input $\mathcal{X}$, $l_{\mathcal{X}}$ denotes a list of predicted probability values for predefined classes in the classification or a numeric value in the regression task, $\text{GNN}$ represents any kind of graph neural networks, $\text{GNN}(\mathcal{G})$ outputs the graph-level representation for $\mathcal{G}$, and $\Phi(\cdot)$ represents a set of task-specific layers, e.g., a binary classification task corresponds to a dense layer followed by a sigmoid function.

\section{Methodology}\label{sec:method}

In this section, we describe the
details of \textsc{GraphRetrieval}. In what follows, we fix an arbitrary set
of training examples $\{ (\mathcal{X}_1, l_1), \ldots,  (\mathcal{X}_n, l_n)  \}$
consisting of pairs $(\mathcal{X}_i, l_i)$
of a graph $\mathcal{X}_i$ and a corresponding prediction $l_i$ (i.e., a class label in the case of classification, or a number in the case of
regression). Furthermore, we fix a GNN given by a set of
message-passing, update, and readout functions. To avoid confusion, we use the superscript to denote retrieved examples. 

\subsection{Initial Training and Indexing}\label{sec:Sec1}

\smallskip 
\noindent\textbf{GNN Training}
The first step in our approach is to train the GNN in a standard supervised way using the given examples. As a result, we obtain a trained GNN model $\mathcal{M}$
with concrete values for each of the parameters of the model; these parameter values will remain fixed in all subsequent steps.

\smallskip 
\noindent\textbf{Indexing of Training Examples} After training, we apply 
$\mathcal{M}$ to each of the  example graphs $\mathcal{X}_1, \ldots \mathcal{X}_n$ and obtain the corresponding graph-level representation vectors $h_{\mathcal{X}_1}, \ldots, h_{\mathcal{X}_n}$ as described in Equation \eqref{eq:graph}.
We then input key-value pairs
$[h_{\mathcal{X}_i}, (\mathcal{X}_i, l_i)]$ for each $1 \leq i \leq n$
to the FAISS retrieval engine~\cite{johnson2019billion}
to construct an index from these key-value pairs. Using a FAISS index will
significantly facilitate the retrieval
of similar graphs at testing time.
It is worth emphasizing that the index
only needs to be constructed once
throughout our entire pipeline. 

\subsection{Training Self-attention Based Adapter} \label{sec:graph_retrieval} 

Given an input $\mathcal{X}$, we compute the corresponding prediction $l$ according to the following steps: First, we apply the trained GNN $\mathcal{M}$ to $\mathcal{X}$ to obtain, in the usual way, an initial prediction $l_{\mathcal{X}}$ and graph-level representation
$h_{\mathcal{X}}$. The main novelty in our approach, however, lies in the following two steps, where
we first retrieve a set of most similar graphs from the index and subsequently exploit self-attention to compute the final prediction $l$. 

\medskip 
\noindent\textbf{Graph Retrieval.} 
In this step, we use the FAISS index to 
find the top$-k$
vectors $\{h_{\mathcal{X}^{1}}, \ldots h_{\mathcal{X}^{k}}\}$ that are most similar
to $h_{\mathcal{X}}$, and to subsequently
retrieve the corresponding example graphs and labels  $(\mathcal{X}^{1}, l^{1}), \ldots (\mathcal{X}^{k}, l^{k})$. 
The value $k$ is considered a hyper-parameter, and we use L2 distance to compute vector similarity scores.\footnote{Recall that, given $d$-dimensional vectors $v^1$ and $v^2$, the $L2$ distance between $v^1$ and $v^2$ is given by $\sqrt{\sum_{i=1}^{d} (v_i^1-v_i^2)^2}$}

\smallskip 
\noindent\textbf{Retrieval Dropout.} Since the index is built from the training dataset, the most similar graph for each input graph is always itself during the training stage. Inspired by~\cite{jiang2021learning} and to reduce overfitting, we also adopt the \textit{retrieval dropout} strategy. Specifically, during the training stage, we will retrieve $k+1$ graphs, in which the most similar graph will be dropped. In the evaluation stage, \textit{retrieval dropout} is disabled, as evaluation graphs are usually not part of the training data.


\medskip 
\noindent\textbf{Self-attention Based Adapter}
We exploit self-attention to compute
the final prediction $l$ for input graph $\mathcal{X}$ as a weighted sum
of the initially predicted label  $l_{\mathcal{X}}$ and the top-$k$ example labels  $l_{i_1}, \ldots l_{i_k}$ with 
respective weights $w_0, \ldots, w_k$.

The self-attention mapping takes as input a 
a \emph{query vector}  and a set of
key-value pairs and yields the weights
$w_0, \ldots, w_k$. In our setting, the query vector 
$\mathbf{q}$ is obtained by
multiplying a 
learnable matrix $\mathbf{W}_1$ to the representation $h_{\mathcal{X}}$
for the input $\mathcal{X}$:
\begin{equation}\label{eq:wq}
     \bm{q} = \bm{W}_{1}\bm{h}_\mathcal{X} ,
\end{equation}
In turn, to obtain the keys,
we pack the $k+1$ graph representations $\{\bm{h}_{\mathcal{X}}, \bm{h}_{\mathcal{X}^{(1)}}, \cdots, \bm{h}_{\mathcal{X}^{(k)}}\} $ as row vectors into a matrix $\bm{H}$, and then multiply $\bm{H}$ with a learnable matrix $\bm{W}_{2}$ to obtain matrix
\begin{equation}\label{eq:wk}
    \bm{K}  = \bm{W}_{2}\bm{H},
\end{equation}

\begin{figure}
    \includegraphics[width=0.5\textwidth]{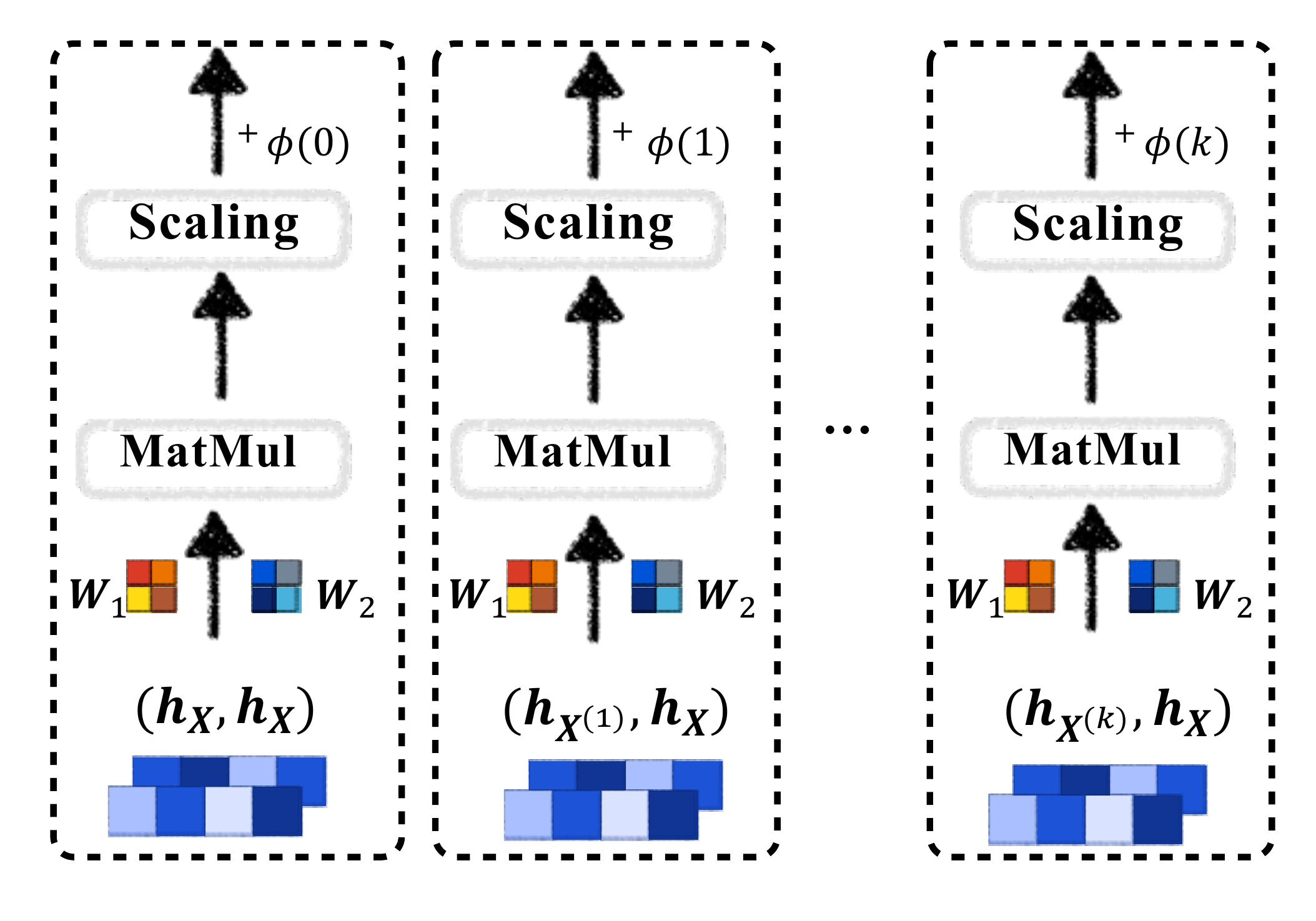}
    \caption{Scheme on Scaled Dot Product Attention.}
    \label{fig:structure}
\end{figure}

Finally, the attention weights
$\bm{Attn} = (w_0, \ldots, w_k)$
are obtained by 
computing the scaled product of $\mathbf{q}$ with the
transpose of $\mathbf{K}$ and adding a $(k+1)$ learnable bias vector $\phi$ (refer to Figure~\ref{fig:structure} for the visual illustration)
\begin{equation}\label{eq:attn}
     \bm{Attn} = \text{softmax}\left(\frac{\bm{q}\bm{K}^T}{\sqrt{d}} + \phi \right),
\end{equation}
where $d$ is the dimension of the query vector $\mathbf{q}$. 
Intuitively, the bias vector $\phi$ allows the model to
encode the ranking between the different retrieved graphs; without it, the model cannot 
distinguish between the different retrieved graphs and the ranking information is lost
as a result.

\smallskip 
\noindent\textbf{Loss Function} Given an input $\mathcal{X}$, the ground-truth label $c$, $l_\mathcal{X}$ and  the top-$k$ example labels $\{l^{1}, \ldots l^{k} \}$, the loss function for the classification task is written as:
\begin{equation}
    \mathcal{L} = - log\left (\bm{Attn}[0]\times l_\mathcal{X}[c] + \sum_{l^{i}=c} \bm{Attn}[i] \right)
\end{equation}
and the loss function for the regression task is written as:
\begin{equation}
    \mathcal{L} = (\sum \bm{Attn} \odot \bm{z} - c)^2
\end{equation}
where $\bm{z} = \left[l_\mathcal{X},l^{1}, \ldots l^{k}\right]$  and $\odot$ is the element-wise product.

\smallskip 
\noindent\textbf{For testing.} 
Given an input $\mathcal{X}$, $l^\mathcal{X}$ and the top-$k$ example labels $\{l^{1}, \ldots l^{k} \}$, the final predicted result for the classification task is: 
$\text{argmax}(L_{\mathcal{X}})$, where $L_{\mathcal{X}} \in \mathbb{R}^{k+1}$ is obtained by adjusting the logits in $l_\mathcal{X}$ -- that is, for  each $c \in [1, \ldots, k]$,
\begin{equation}
    L_{\mathcal{X}}[c] = \bm{Attn}[0]\times l_\mathcal{X}[c] + \sum_{l^{i}=c} \bm{Attn}[i]
\end{equation}
and the final predicted result for the regression task is:
$\sum \bm{Attn} \odot \bm{z}$, where $\bm{z} = \left[l_\mathcal{X},l^{1}, \ldots l^{k}\right]$  and $\odot$ is the element-wise (Hadamard) product.

\subsection{Two-phase Model Training}\label{sec:train}

{
\setlength{\textfloatsep}{0pt}
\begin{algorithm}[ht!] 
\caption{Pipeline of Two-phase Model Training} 
\label{alg:pipeline} 
\begin{algorithmic}[1] 
\REQUIRE
Training and validation datasets; randomly initialised GNN and Adapter; training epochs $m_1$ for GNN and $m_2$ for the adapter; the number of retrieved graphs $k$.
\FOR{$i=1$ {\bfseries to} $m_1$}
\STATE Train the GNN model over the training
dataset. Optimal GNN parameters of a trained model $\mathcal{M}$ are saved by evaluating the validation dataset.
\ENDFOR
\STATE Construct FAISS index.
\FOR{$i=1$ {\bfseries to} $m_2$}
\STATE Train the Adapter over  the train set using the index and the fixed model $\mathcal{M}$, where the \textit{retrieval dropout} strategy is applied in this stage. The optimal
parameters of the Adapter are saved by evaluating the validation dataset.
\ENDFOR
\RETURN $\mathcal{M}$ and Adapter defined by $\bm{W}_1$, $\bm{W}_2$, $\bm{b}$.
\end{algorithmic} 
\end{algorithm}
}

The overall training pipeline
is illustrated in~ Algorithm~\ref{alg:pipeline}.
%
Lines 1--4 correspond to the initial training of
the GNN model and the construction of the index. 
In Lines 5--7 we train the adapter module and
optimise the values in the learnable matrices $\bm{W}_1$ and $\bm{W}_2$ and the bias vector 
$\bm{b}$ described in Section \ref{sec:graph_retrieval}, where the GNN model $\mathcal{M}$ is fixed.
Finally, when learning $\bm{W}_1$ and $\bm{W}_2$ from the training data using our entire pipeline, it is worth noticing that the most similar graph of an example graph $\mathcal{X}$ will be itself, so we will always drop it and consider only the other retrieved graphs. 
\section{Experiments} \label{sec:experiments}
In this section, we report empirical evaluations of \textsc{GraphRetrieval} for 13 graph property prediction datasets. We show that retrieval-enhanced models yield excellent performance gains when compared with baseline GNN models\footnote{Our codes are available in \url{https://github.com/wdimmy/GraphRetrieval}.}.

\subsection{Benchmark Dataset \label{sec:benchmark_dataset}}
 We used a collection of $13$ datasets consisting
 of $8$ small-scale molecule datasets~\cite{hu2020open}, $2$ computer vision datasets~\cite{dwivedi2020benchmarking}, $1$ large-scale therapeutics-related dataset
 (USPTO~\cite{lowe2012extraction}) and
$2$ large-scale quantum chemistry datasets
(PCQM4M and PCQM4Mv2~\cite{hu2021ogb}). Detailed information on these Benchmark datasets is summarized in Table~\ref{tab:summary}. 

\begin{table}
    \centering
    \scriptsize
    \setlength{\tabcolsep}{5.2pt}
    \renewcommand{\arraystretch}{1.7}
    \caption{Statistics of datasets and their evaluation metrics. MC and BC denote multi-class and binary classification, respectively. For PCQM4M and PCQM4Mv2, their testing datasets are not publicly available. We report the best result on the development sets.
    }
    \label{tab:summary}
    \begin{tabular}{lrrrrr}
      \toprule
         \textbf{Dataset}& \textbf{Train} & \textbf{Valid} & \textbf{Test} & \textbf{Task} &  \textbf{Metric} \\
         \hline 
         BBBP & 1631 & 204 & 204 & BC & ROC\_AUC \\ 
         Tox21 & 6264 & 783 & 784 & BC & ROC\_AUC \\
         ToxCast & 6860 & 858 & 858 & BC & ROC\_AUC \\
         SIDER & 1141 & 143 & 143 & BC & ROC\_AUC \\
         ClinTox& 1181 & 148 & 148 & BC & ROC\_AUC \\
         MUV& 74469 & 9309 & 9309 & BC & ROC\_AUC \\
         HIV & 32901 & 4113 & 4113 & BC & ROC\_AUC \\
         BACE & 1210 & 151 & 152 & BC & ROC\_AUC \\
         \hline 
         PCQM4M& 3,045,360 & 380,670 & -- & Regression & MAE  \\
         PCQM4Mv2 & 3,378,606 & 73,545  & --  & Regression & MAE \\ 
         \hline 
         MNIST & 55000 & 5000 & 10000 & MC & Accuracy  \\
         CIFAR10 & 45000 & 5000 & 10000 &  MC & Accuracy \\
         \hline 
         USPTO & 505,259 & 72,180 & 144,360 & MC & Accuracy \\ 
      \bottomrule 
    \end{tabular}
\end{table}

\begin{table}
    \centering
    \caption{Experimental results of different methods on the four different testing datasets. The number in the bracket of \emph{Majority-voting} method represents the retrieved number. 
    }
    \renewcommand{\arraystretch}{1.1}
    \label{tab:q2}
    \begin{tabular}{lrrrr}
    \toprule
        & \textbf{CIFAR10} & \textbf{MNIST} & \textbf{HIV} &  \textbf{USPTO}\\
      \hline 
       GCN & \textbf{64.7} &  \textbf{96.0} & \textbf{78.6}  & \textbf{33.5} \\
      \hline 
      Retrieval & 49.1 & 85.3 & 54.1 & 24.3 \\
      Majority-voting~(5) & 51.7 & 87.1  &   57.3 & 25.7 \\
      Majority-voting~(10) & 51.9 & 88.9 &   58.1 & 26.9 \\
      Majority-voting~(20) & 48.7 & 84.2 &   56.3 & 25.2\\
       \midrule 
       \midrule 
     GIN & \textbf{65.6} &  \textbf{96.8} & \textbf{77.0}  & \textbf{38.6} \\
      \hline 
       Retrieval& 51.1 & 86.2 & 55.8 & 26.5 \\
       Majority-voting~(5) & 51.1 & 87.9  &   56.9& 27.7 \\
     Majority-voting~(10) & 52.9 & 89.4 &   58.9& 28.2 \\
       Majority-voting~(20) & 52.7 & 88.6 &   57.3 & 26.3\\
       \midrule 
       \midrule 
      PNA & \textbf{68.7} &  \textbf{97.2} & \textbf{77.0}  & \textbf{35.2} \\
      \hline 
      Retrieval & 51.8 & 86.2 & 56.1 & 26.8 \\
      Majority-voting~(5) & 52.1 & 87.9  &   57.6 & 27.4 \\
      Majority-voting~(10) & 52.9 & 88.9 &   58.2 & 27.4 \\
      Majority-voting~(20) & 52.2 & 87.4 &   57.8 & 26.4\\
    \bottomrule
    \end{tabular}
\end{table}

\subsection{Baseline Models \label{sec:baselinne_model}}
We adopt three widely-used GNN baselines: Graph Convolutional Networks~(GCN) \cite{kipf2016semi},  
Graph Isomorphism Networks (GIN)~\cite{xu2018powerful},
and Principal Neighbourhood Aggregation~(PNA)~\cite{corso2020principal}. For these baselines, we take the default values of the hyperparameters specified in the publicly-released GitHub repositories. Besides, we adopt two purely retrieval-based approaches:
\begin{itemize}
    \item \emph{Retrieval}, which retrieves the most similar graph from the FAISS index whose label is as our predicted answer. Note that when evaluating using the ROC-AUC metric, we instead use the corresponding similarity score returned by FAISS. 
    \item \emph{Majority-voting}, which retrieves some number of the most similar graphs from the FAISS index whose labels will be ranked based on their frequency, and then we choose the most frequent label as our predicted answer. Note that when evaluating using the ROC-AUC metric, we need to use the similarity score instead of the predicted label. Unlike the \emph{Retrieval} strategy, in the \emph{Majority-voting} strategy, the most frequent label may correspond to multiple instances, leading to multiple similarity scores. In such cases, we select the highest similarity score among them.
\end{itemize}

\begin{table}
    \centering
    \renewcommand{\arraystretch}{1}
    \caption{Experimental results of Self-attention VS Averaging.}
    \label{tab:q3}
    \begin{tabular}{llrrrr}
    \toprule
      \multicolumn{2}{c}{-}   & \textbf{CIFAR10} & \textbf{MNIST} & \textbf{HIV} &  \textbf{USPTO} \\
      \hline 
      \multirow{2}{*}{GCN} & Self-attention & \textbf{67.6} &\textbf{97.2} &  \textbf{80.8} & \textbf{39.2} \\
      & Averaging &  57.7 & 92.1  &  60.7 & 30.6 \\
    \midrule 

      \multirow{2}{*}{GIN} & Self-attention & \textbf{68.2} &\textbf{97.9} &  \textbf{79.5} & \textbf{42.8} \\
      & Averaging &  58.6 & 93.2  &  59.3 & 33.1 \\
     \midrule 
      \multirow{2}{*}{PNA} & Self-attention & \textbf{72.3} &\textbf{98.2} &  \textbf{78.0} & \textbf{41.6} \\
      & Averaging &  58.7 & 92.9  &  58.2 & 32.8 \\
    \bottomrule
    \end{tabular}
\end{table}

\subsection{Training Details} \label{sec:implementation}
$\textsc{GraphRetrieval}$ is an agnostic to the choice of graph neural networks, and can be applied to enhance a wide range of GNN models. Training is conducted as described in 
Algorithm \ref{alg:pipeline}, where $m1=300$, $m2=200$ and $k=3$\footnote{Different numbers of retrieved examples might have different effects for our $\textsc{GraphRetrieval}$ framework, and further improvements might be achieved by choosing the number of retrieved examples in a more fine-grained way~(e.g., differ across different datasets).  In this paper, we choose $k=3$ according to our preliminary experiments and we leave further exploration of the effects of the number of retrieved examples for future work.}.
We use the Adam optimizer~\cite{DBLP:journals/corr/KingmaB14} with an initial learning rate $0.01$ and decay the learning rate by $0.5$ every $50$ epochs for all the GNN baselines. Batch normalization~\cite{ioffe2015batch} is applied on every hidden layer. Some hyperparameters for the three GNN baseline models are shown in 
Phase 2 of the training (Lines  5--7) is repeated $5$ times
with different initialisation seeds for
$\bm{W}_1$, $\bm{W}_2$, and $\bm{b}$; we report the mean and standard deviation of the values obtained in these runs.

\subsection{Exploratory Experiments}

\noindent\textbf{Predict with only graph retrieval.}  We compare \emph{Retrieval} and \emph{Majority-voting} methods against well trained graph neural network models on four datasets: CIFAR10, MNIST, PCQM4M and PCQM4Mv2. In particular, for the Majority-voting method, we try three different retrieved numbers to see their performance on four different datasets. The experimental results are shown in Table~\ref{tab:q2}.

From Table~\ref{tab:q2}, we can see that both \emph{Retrieval} and \emph{Majority-voting} method greatly underperform the three well-trained models. For the four datasets with different baseline GNN models, \emph{Retrieval} methods have about $10\% \sim 15\%$ accuracy loss when compared with  the well-trained GNN model. For the Majority-voting method, we can notice that as we increase the number of retrieved graphs from $10$ to $20$, the accuracy decreases on all four datasets. One straightforward reason might be that as the number of retrieved results increases, the number of noisy labels will increase accordingly. 

\smallskip 

\begin{table*}[t]
    \centering
    \setlength{\tabcolsep}{3pt}
    \renewcommand{\arraystretch}{1.4}
    \caption{Test ROC-AUC ($\%$) performance on $8$ molecular prediction Benchmark datasets.}
    \label{tab:molecules}
    \begin{tabular}{llrrrrrrrr}
    \toprule
         &  \textbf{BBBP} & \textbf{Tox21} & \textbf{ToxCast} & \textbf{SIDER} & \textbf{ClinTox} & \textbf{MUV} & \textbf{HIV} & \textbf{BACE} & \textbf{Average}\\
      \hline 
        GCN & 70.0($\pm$0.3) & 73.9($\pm$1.8) & 64.0($\pm$2.3) & 58.6($\pm$2.1) & 92.2($\pm$2.7) & 75.9($\pm$2.0) & 78.6($\pm$0.3) & 81.9($\pm$0.9) & 74.4 \\
       GIN & 70.5($\pm$0.7) & 74.9($\pm$0.2) & 64.5($\pm$1.0) & 58.7($\pm$2.1) & 87.0($\pm$3.0) & 76.8($\pm$1.8) & 77.0($\pm$0.2) & 74.8($\pm$0.7) & 73.0 \\
       PNA & 70.0($\pm$1.9) & 73.0($\pm$0.3) & 62.0($\pm$1.9) & 58.0($\pm$0.2) & 87.0($\pm$2.3) & 71.0($\pm$3.5) & 77.0($\pm$0.6) & 72.0($\pm$0.5) & 71.2 \\
        \hline 
        \hline 
      Retrieval-enhanced GCN & 71.0($\pm$1.2) & 75.9($\pm$1.0) & 65.4($\pm$0.2) & 60.8($\pm$0.3) & 93.2($\pm$2.5) & 77.7($\pm$0.8) & 80.8($\pm$1.0) & 84.1($\pm$1.8) & \textbf{76.1} \\
      Retrieval-enhanced GIN & 71.9($\pm$2.5) & 77.3($\pm$0.2) & 67.4($\pm$1.6) & 61.5($\pm$1.03) & 89.2($\pm$2.6) & 78.4($\pm$1.7) & 79.5($\pm$0.9) & 81.2($\pm$0.7) & 75.8\\
     Retrieval-enhanced PNA & 71.5($\pm$1.4) & 75.4($\pm$1.8) & 63.1($\pm$0.3) & 59.8($\pm$0.4) & 89.0($\pm$2.0) & 72.9($\pm$1.0) & 78.0($\pm$0.9) & 79.9($\pm$1.9) & 73.7\\
    \bottomrule
    \end{tabular}
\end{table*}



\noindent\textbf{Self-attention vs. Averaging.} One intuition about using the self-attention is that
we hope the \textit{Self-attention Adapter} module can learn to weigh retrieved examples differently such that the more “important” retrieved results correspond to the higher weights. Here, to demonstrate the necessity and the effectiveness of using the self-attention mechanism in our framework, we compare a self-attention setting against an averaging setting. To be more specific, each label will be assigned an equal weight without distinguishing their degree of importance to our prediction based on Equation~\ref{eq:attn}.

From Table~\ref{tab:q3}, we can observe that incorporating the self-attention mechanism into our framework significantly outperforms using the averaging scheme. For example, for the  USPTO dataset, using the self-attention mechanism achieves about $22\%$ accuracy improvement over the averaging scheme. One possible reason is that the retrieved results are not always useful, so the model should not assign the same weight to the retrieved labels and the label obtained by the well-trained model in computing the final prediction. Therefore, adopting self-attention to model the relevance between the target graph and the retrieved graphs is considered to be an effective way to make use of the retrieved results while avoiding sabotaging the performance of the well-trained model.

\begin{table}
    \centering
    \small 
    \caption{MAE Performance on PCQM4M and PCQM4Mv2.}
    \label{tab:pcqm4m}
    \begin{tabular}{lrr}
    \toprule
       & \textbf{PCQM4M} & \textbf{PCQM4Mv2}\\
      \hline 
      GCN & 0.171&  0.139 \\
      GIN & 0.157 &  0.123 \\
      \hline 
      \hline 
      Retrieval-enhanced GCN &0.160 & 0.132\\
      Retrieval-enhanced GIN &\textbf{0.142}&\textbf{0.115}\\
    \bottomrule
    \end{tabular}
\end{table}

\smallskip 

\noindent\textbf{Retrieval Dropout or Not?} We investigated the effect of retrieval dropout in our proposed \textsc{GraphRetrieval} framework. Our ablation studies on various datasets demonstrate that incorporating retrieval dropout consistently leads to improved performance compared to not using the strategy. In particular, we observed that the gain is less significant in the two image datasets compared to the molecular dataset (e.g., only a $0.05\%$ decrease in accuracy for MNIST but a $0.4\%$ performance drop in HIV datasets). One possible explanation is that the difference between training and inference in the image datasets is not as pronounced as in the molecular datasets. Overall, our experimental results suggest that this training strategy enhances the generalization ability of \textsc{GraphRetrieval}.

\subsection{Main Results
\label{sec:main_results}}

We can observe that the use of \textsc{GraphRetrieval}
significantly enhances the performance of each
of the baselines, which demonstrates the effectiveness of our approach. We next discuss each of these results in further detail.

\begin{table}
    \centering
    \small 
    \caption{Classification accuracy on CIFAR10, MNIST and  USPTO.}
    \label{tab:images}
    \begin{tabular}{lrrr}
    \toprule
      -   & \textbf{CIFAR10} & \textbf{MNIST} & \textbf{USPTO}\\
      \hline 
      GCN & 64.7 &  96.0 & 33.5\\
      GIN & 65.6 &  96.8 & 38.6 \\
      PNA & 68.7 &  97.2 & 35.2 \\
      \hline 
      \hline 
      Retrieval-enhanced GCN & 67.6 &97.2 & 39.2\\
      Retrieval-enhanced GIN & 68.2&97.9 & \textbf{42.8} \\
       Retrieval-enhanced PNA & \textbf{72.3}& \textbf{98.2} & 41.6  \\
    \bottomrule
    \end{tabular}
\end{table}

\begin{figure}[ht]
    \centering
    \includegraphics[width=0.47\textwidth]{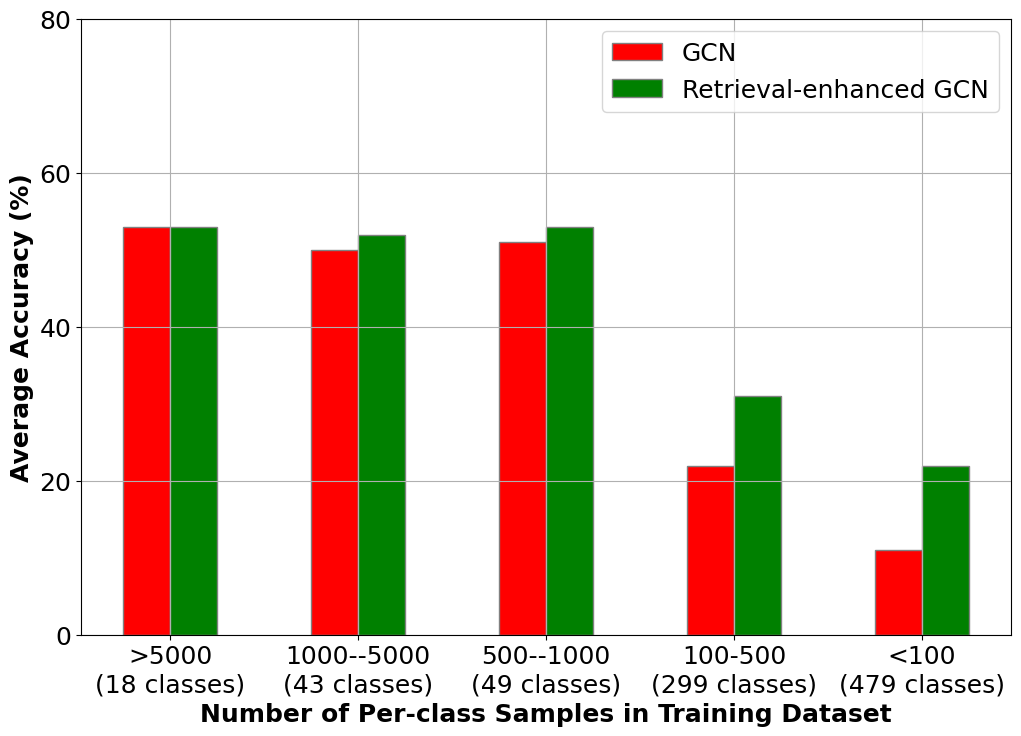}
    \caption{Dissected experiments on USPTO dataset. We classify the training datasets into five groups according to the number of associated samples of each class in the training dataset. 
    }
    \label{fig:longtail}
\end{figure}

\smallskip
\noindent\textbf{Molecule Datasets.} These datasets are highly imbalanced since the label assigned to most samples is $0$;
The task is to predict the target molecular properties as accurately as possible, where the molecular properties are cast as binary labels, e.g, whether a molecule inhibits HIV virus replication or not.
Our results are  summarised in Table~\ref{tab:molecules}.
 First, we observe that \textsc{GraphRetrieval} improves the classification performance of each of the baselines for all eight molecule datasets; furthermore, the improvements are consistent across all three baselines, which demonstrates that our approach is effective even on highly imbalanced datasets. Second, since these datasets use scaffold spitting, the substructures of the evaluation graphs are rarely reflected in the training graphs; it is therefore encouraging to observe that our approach is helpful even in such a challenging setting.

\begin{table*}[ht!]
    \centering
    \setlength\tabcolsep{8pt}
    \begin{tabular}{llll}
        \toprule
        Model & Input & Predicted Probability  & Retrieval \\ 
        \hline
        PNA &  \raisebox{-.4\totalheight}{\includegraphics[width=1.4cm,height=1cm]{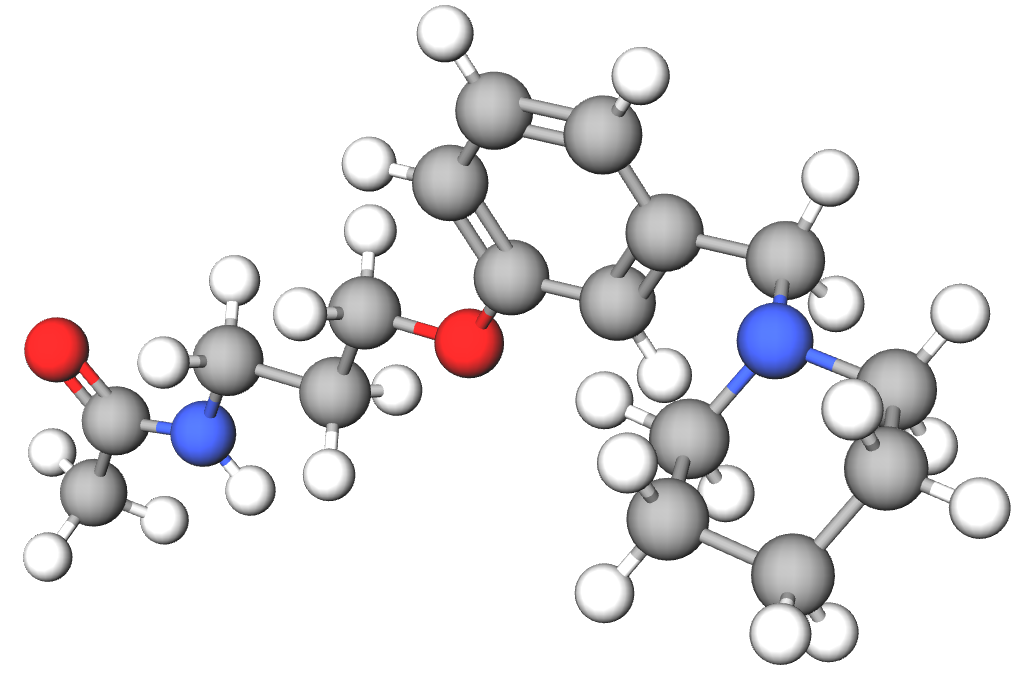}}~(label=1)  & $P_{label=1} = 0.32$  & No Retrieval   \\
        Retrieval-enhanced PNA & \raisebox{-.4\totalheight}{\includegraphics[width=1.4cm,height=1cm]{figures/1.png}}~(label=1)  & $P_{label=1} = 0.81$ &  (\raisebox{-.4\totalheight}{\includegraphics[width=1.4cm,height=1cm]{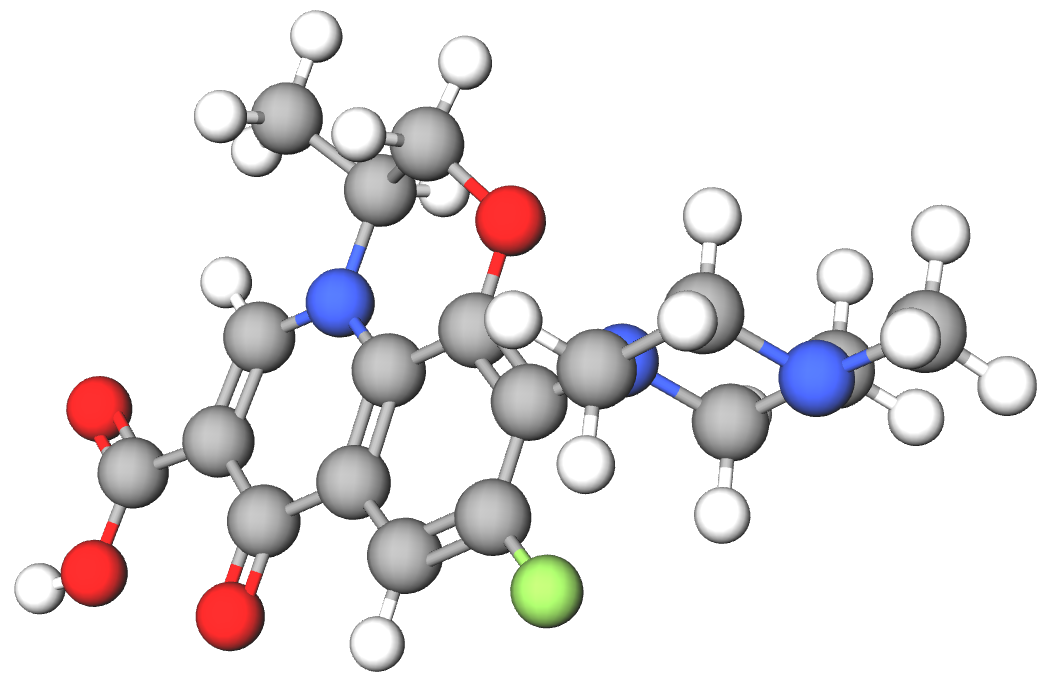}},\textcolor{green}{1})   
        (\raisebox{-.4\totalheight}{\includegraphics[width=1.4cm,height=1cm]{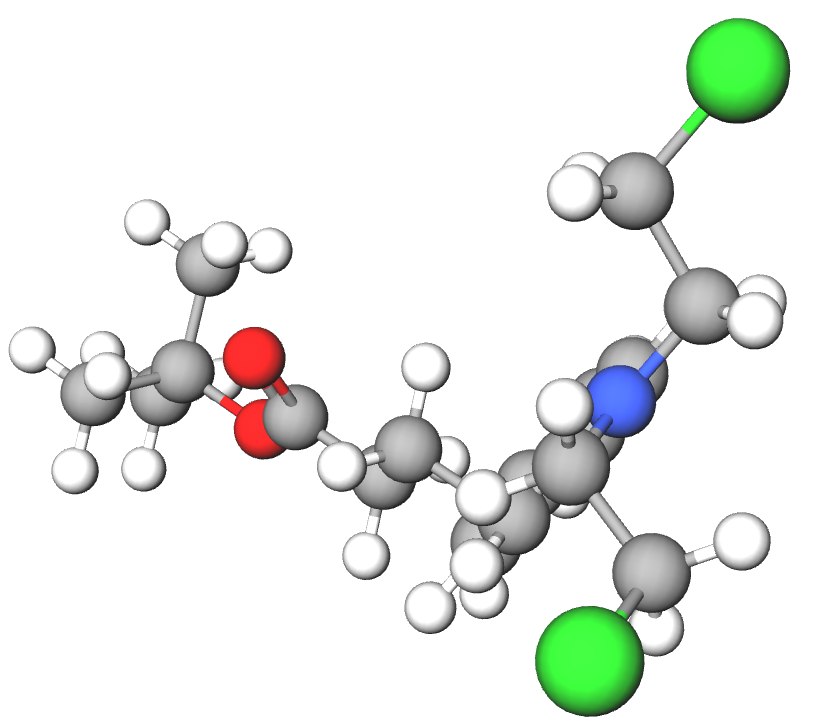}},\textcolor{green}{1})
        (\raisebox{-.4\totalheight}{\includegraphics[width=1.4cm,height=1cm]{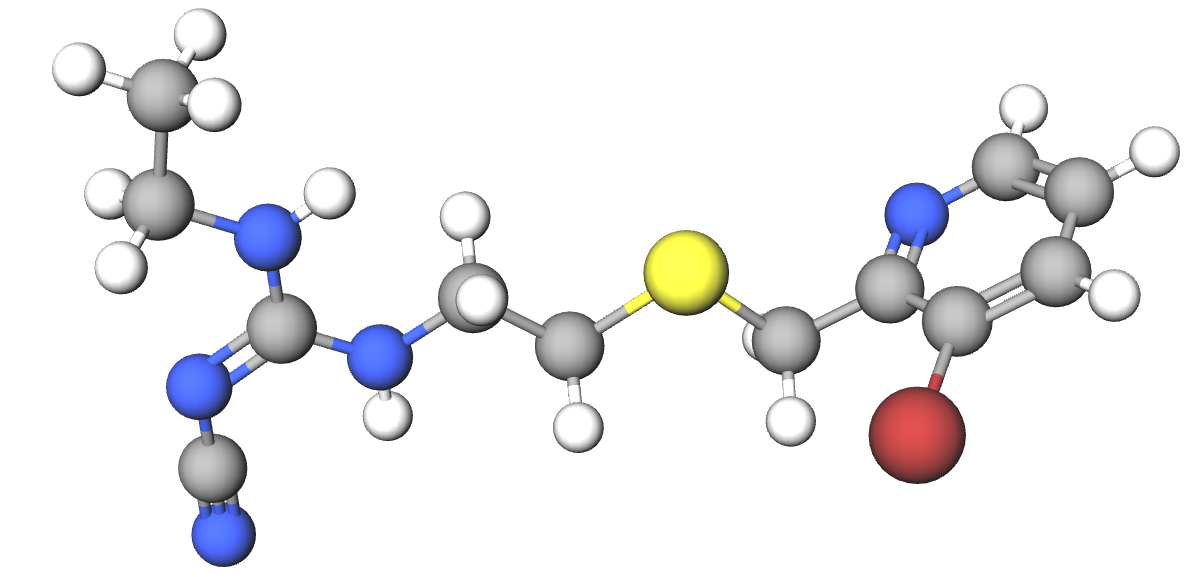}},\textcolor{red}{0}) \\
        \midrule
        \midrule
         PNA &  \raisebox{-.4\totalheight}{\includegraphics[width=1.4cm,height=1cm]{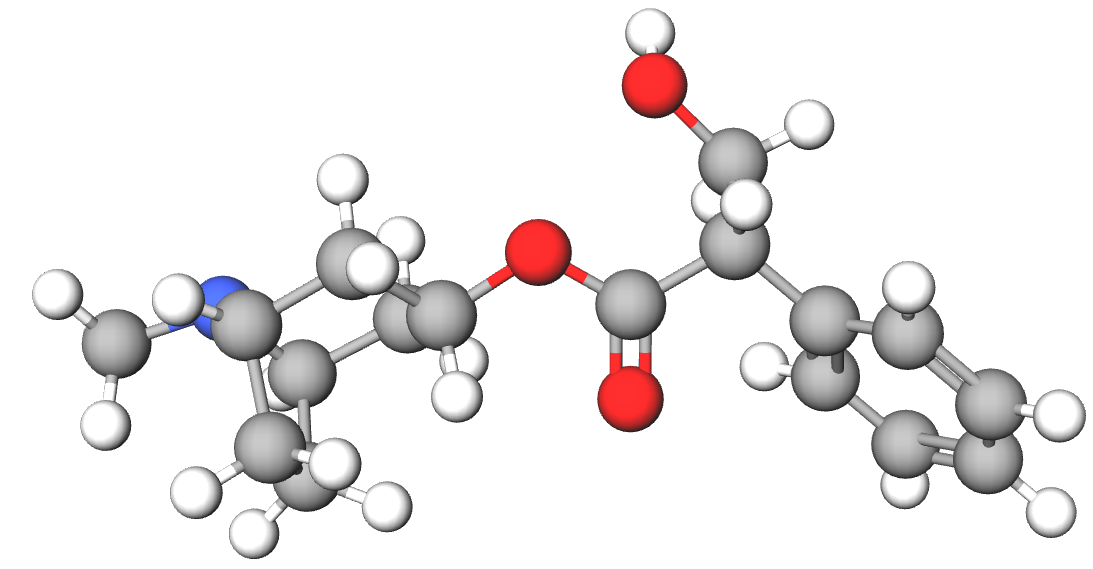}}~(label=1)  & $P_{label=1} = 0.79$  & No Retrieval   \\
        Retrieval-enhanced PNA & \raisebox{-.4\totalheight}{\includegraphics[width=1.4cm,height=1cm]{figures/4.png}}~(label=1)  & $P_{label=1} = 0.74$ &  (\raisebox{-.4\totalheight}{\includegraphics[width=1.4cm,height=1cm]{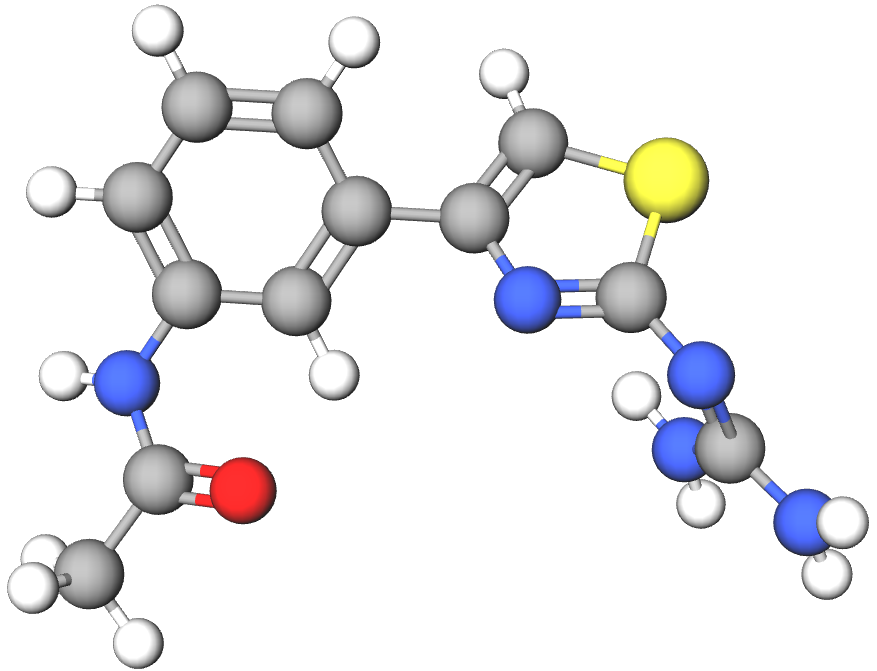}},\textcolor{red}{0})   
        (\raisebox{-.4\totalheight}{\includegraphics[width=1.4cm,height=1cm]{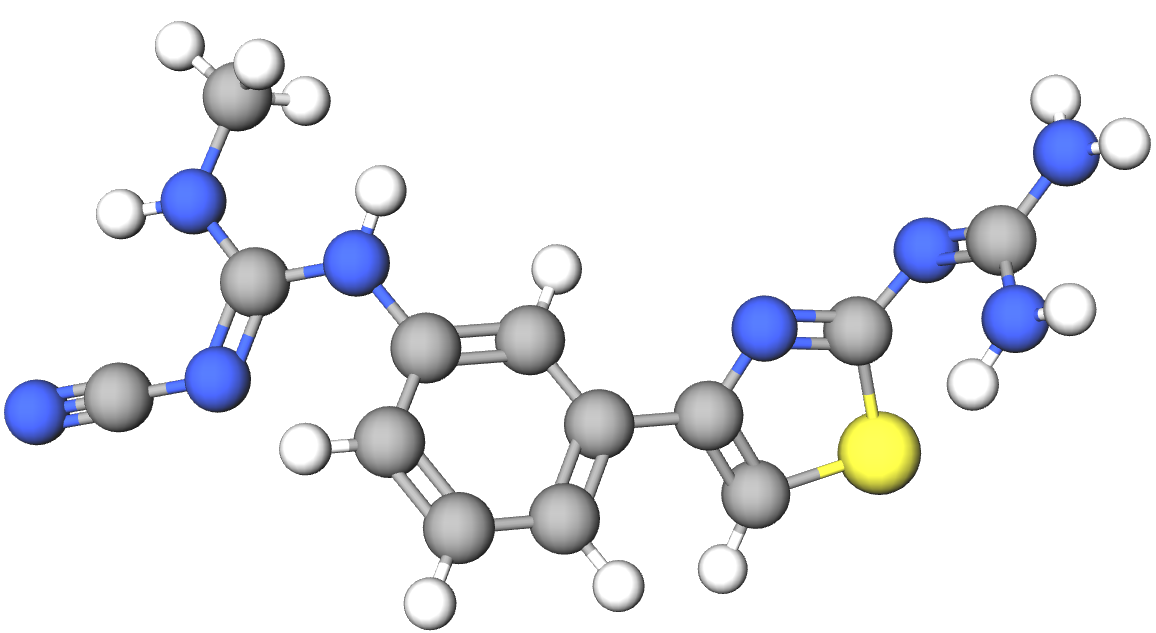}},\textcolor{red}{0})
        (\raisebox{-.4\totalheight}{\includegraphics[width=1.4cm,height=1cm]{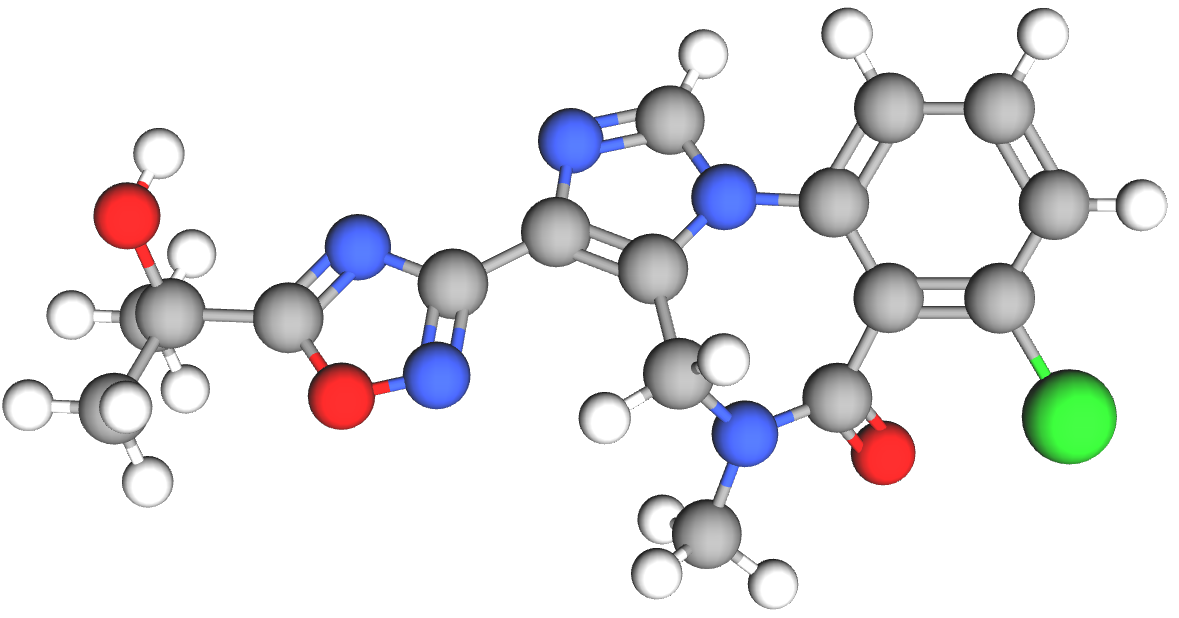}},\textcolor{red}{0}) \\
        \bottomrule
    \end{tabular}

\bigskip

    \caption{Two cherry-picked examples showing how \textsc{GraphRetrieval} “absorbs” useful information from
retrieved graphs and “ignores” noise introduced by irrelevant graphs, respectively. In the first example, it assigns a much higher probability (0.81) to the correct label, ``1'', compared to PNA. In the second example when retrieved graphs are considered to be noisy information, both retrieval-enhanced PNA and PNA assign a similar probability~(0.74 and 0.79, respectively) to the correct label.}
    \label{my-label}
\end{table*}

\noindent\textbf{MNIST, CIFAR10 and USPTO.} Our results
on these datasets are summarised in Table~\ref{tab:images}. First, in contrast to the molecule datasets, where graphs are highly diverse and exhibit heterogeneous topologies, the two computer vision datasets (MNIST and CIFAR10) rely on grid-structured graphs with a fixed topology~\cite{corso2020principal}. From Table \ref{tab:images}, we can observe that these results are  consistent with those obtained for the molecule datasets, and hence our
approach is also effective for computer vision problems. Besides, we experiment with another large-scale multi-classification dataset (USPTO). From Table~\ref{tab:images}, we can notice that the performance improvement on USPTO is much larger than the other two image datasets. One possible reason is that different from MNIST and CIFAR10, USPTO has more classes (888) and the distribution of instances for each class is long-tailed.

To gain more insights into how the retrieval-enhancement help improve the overall performance of baseline models, we trace the model's performance in each class. We first classify $888$ classes into five groups according to the number of samples appearing in the training datasets. Then, we report the accuracy of each class in the testing dataset. To avoid the class imbalance issue, we take a subset of the original testing, such that each class has the same number of samples needed to be predicted. From Figure~\ref{fig:longtail}, we observe that the improvement achieved by the retrieval enhancement in $100-500$ and $<100$ groups is much larger than the other three groups, showing that the retrieval enhancement is a promising remedy for increasing the accuracy of those ``long-tailed'' classes associated with only a few samples in the training dataset. 

\smallskip 

\noindent\textbf{PCQM4M and PCQM4Mv2.} These are  two
large-scale datasets consisting of millions of graphs. The task is graph regression: predicting the HOMO-LUMO energy gap in electronvolt given 2D molecular graphs. From Table~\ref{tab:pcqm4m}, we found that \textsc{GraphRetrieval} brings significant performance improvements on both datasets, which suggests that our approach is also beneficial for regression tasks. In particular, as shown in Figure~\ref{fig:example}(c),  when the training dataset shows a long-tailed value distribution, we can see that \textsc{GraphRetrieval} can significantly increase the performance  for predicting  those long-tailed when compared with baseline models. These results are consistent with the results in the classification task (USPTO) wherein many labels are associated with only a few samples, indirectly showing that our proposed approach  exhibits substantial generality for improving the model's performance in different tasks.

Overall, our results have demonstrated that \textsc{GraphRetrieval} can bring substantial gains to GNN models. Such results are very promising because it first validates the feasibility of retrieval-enhanced approaches in the world of graph neural networks. Besides, another encouraging point is that the improvement does not come at the significant cost of increasing the model size as other models. 

\subsection{Case Study}\label{sec:case-study}
To conclude this section, we cherry-pick two examples of Retrieval-enhanced PNA and PNA predicting the molecular property.  In the first example, ``1'' is the correct label of the given input, and Retrieval-enhanced PNA gives the label=1 a much high probability compared to the PNA model. Since Retrieval-enhanced PNA manages to retrieve some similar graphs with possibly related labels, the probability solely calculated based on the input graph will be rectified. In the second example, the labels of the three retrieved graphs are all different from our ground-truth label, so they are considered to be noisy information. However, both Retrieval-enhanced PNA and PNA output a similar probability for the correct label, which denotes that Retrieval-enhanced PNA is resistant to  noisy information. One possible reason is that  our proposed \textsc{GraphRetrieval} scheme will assign lower weights to those noisy labels by the self-attention mechanism. Overall, the case study shows that retrieved graphs can facilitate graph property prediction and the retrieval-enhanced approach is applied to graph neural networks. 
\section{Conclusion and Future Work} \label{sec:conclusion}

We have proposed \textsc{GraphRetrieval}, a  model-agnostic approach for 
improving the performance of GNN baselines on graph classification and regression tasks. Our empirical results clearly show that \textsc{GraphRetrieval} consistently improves the performance of existing GNN architectures on a very diverse range of benchmarks. 
We believe that our results open the door for the further development of sophisticated methods for explicitly exploiting the training data when making predictions on graphs.

Our research opens a number of interesting avenues for future work. First, it would be interesting to extend our approach to tackle node-level classification tasks. 
Furthermore, our approach is rather general and flexible, and we believe that it can be applied to advanced methods based on 3D geometric modeling
~\cite{schutt2017schnet,schutt2021equivariant} as well as to a broad range of additional tasks~\cite{gilmer2017neural,townshend2020atom3d}.

\section*{Acknowledgments}
This work was supported in whole or in part by
the EPSRC projects OASIS (EP/S032347/1), ConCuR (EP/V050869/1)
and UK FIRES (EP/S019111/1),
the SIRIUS Centre for Scalable Data Access, and
Samsung Research UK. 


\bibliography{ecai}
\end{document}